\documentclass{esannV2}
\usepackage[pdftex]{graphicx}
\usepackage[latin1]{inputenc}
\usepackage{amssymb,amsmath,array}

\newcommand{\R}[1]{\ensuremath{\mathbb{R}^{#1}}}
\newcommand{\itemSet}{\Omega}
\newcommand{\partition}{\mathcal{P}}
\newcommand{\error}{E}
\newcommand{\cluster}{C}
\newcommand{\sumDist}{S}
\newcommand{\reductionFactor}{\alpha}

\begin{document}
\title{Dissimilarity Clustering by Hierarchical Multi-Level Refinement}

\author{Brieuc Conan-Guez$^1$ and Fabrice Rossi$^2$
%
%
\vspace{.3cm}\\
%
1-  LITA EA 3097, Université Paul Verlaine-Metz\\
Île du Saulcy 57045 Metz cedex 1, France
%
\vspace{.1cm}\\
2- SAMM EA 4543, Université Paris 1 Panthéon-Sorbonne\\
90, rue de Tolbiac, 75634 Paris cedex 13, France\\
}

\maketitle

\begin{abstract}
  We introduce in this paper a new way of optimizing the natural extension of
  the quantization error using in k-means clustering to dissimilarity
  data. The proposed method is based on hierarchical clustering analysis
  combined with multi-level heuristic refinement. The method is
  computationally efficient and achieves better quantization errors than the
  relational k-means.
\end{abstract}

\section{Introduction}
Non vector data arise in numerous real world applications for which 
complex object representations are needed, ranging from variable size strings
to tree and graph structure observations. Such data can be analysed by means
of dissimilarity measures: one works only on the square matrix of all pairwise
dissimilarities between the observations. Numerous data analysis methods have
been adapted or directly defined to handle dissimilarity matrices, from
$k$-nearest neighbors in a supervised context to self organizing maps in the
unsupervised one. 

A traditional way of clustering proximity data is to apply the usual
hierarchical clustering analysis (HCA) technique 
(see e.g. \cite{Anderbeg73ClusterAnalysis}). 
However, this elementary solution suffers from three potential difficulties. Firstly, classical linkage
criteria for HCA such as single, average or complete, are not very
satisfactory as they do not derive from a global quality criterion for the
obtained partition. Secondly, naive implementations of the HCA scale in
$O(N^3)$ (where $N$ is the number of observations) which is unacceptable for
large datasets. Thirdly, as a greedy technique, HCA leads to suboptimal
partitions.  We address in this paper those three limitations in order to
derive an efficient alternative to prototype based dissimilarity clustering
techniques such as the relational k-means
\cite{RossiEtAlFastRelationalWSOM2007}. Following \cite{Batagelj1988}, we
introduce a linkage criterion which corresponds to the natural extension of
the quantization error to dissimilarity data. Then we use the HCA algorithm
from \cite{Mullner11ModernHierarchical} to obtain a fast
implementation. Finally, we apply multi-level refinement heuristics coming
from graph clustering to improve the suboptimal partitions.

\section{Hierarchical Clustering Analysis}

\subsection{Dissimilarity matrix and error measure}\label{sectionErrorMeasure}
We consider the general case of a set $\itemSet$ equipped with a dissimilarity
measure. This is a map from $\itemSet \times \itemSet \rightarrow \R{+}$,
which is reflexive and symmetric, i.e., $d(i,i)=0$ and $d(i,j)=d(j,i)$ for all
$i, j \in \itemSet$. We consider a partition of $\itemSet$ in $c$ distinct
clusters: $\partition = \{ \cluster_1,\ldots, \cluster_c \}$.  Denoting
$\sumDist_{\cluster_p,\cluster_q} = \sum_{i \in \cluster_p, j \in \cluster_q}
d(i,j)$, and $\sumDist_{\cluster_p} = \sumDist_{\cluster_p, \cluster_p}$, the
quality of a partition is measured thanks to the classical error measure
\cite{HofmannBuhmann1997TPAMI}:
\[
\error(\partition) = \sum_{\cluster_k \in \partition} \frac{1}{|\cluster_k|}
\sum_{i,j \in \cluster_k} d(i,j) = \sum_{\cluster_k \in \partition}
\frac{\sumDist_{\cluster_k}}{|\cluster_k|},
\]
where $|\cluster_k|$ is the size of cluster $\cluster_k$. The
normalization term $\frac{1}{|\cluster_k|}$ can be seen as a way to favor
partitions with clusters of intermediate sizes. It arises naturally in the
case of vector data: indeed, as shown in e.g. \cite{HofmannBuhmann1997TPAMI},
$\error(\partition)$ is equal to twice the quantization error when the
dissimilarity is the square of the Euclidean distance between observations.

Merging two clusters $\cluster_p$ and $\cluster_q$ of the partition
$\partition$ leads to a new partition $\partition'$ with an error increased by:
\[
\Delta \error_{p,q} =   \frac{\sumDist_{\cluster_p \cup
    \cluster_q}}{|\cluster_p|+|\cluster_q|}  -
\frac{\sumDist_{\cluster_p}}{|\cluster_p|}  -
\frac{\sumDist_{\cluster_q}}{|\cluster_q|}.
\]
This increase can be used as a linkage criterion for a hierarchical clustering
analysis (HCA). The derivation of this criterion generalizes Ward's derivation
for the squared Euclidean distance (see \cite{Batagelj1988}). 

In a naive implementation of the HCA, each step of the algorithm proceeds by
merging the pair of clusters, $\cluster_p$ and $\cluster_q$, which leads to
the smallest increase of the error $\error$ (that is the closest pair
according to the linkage).  If $N$ denotes the size of $\itemSet$, the
hierarchy is composed of a sequence of $N$ partitions (from the finest to the
coarsest).  The algorithm complexity is $O(N^3)$, using the fact that
computation of sums $\sumDist_{\cluster_p \cup \cluster_q}$ can be done
incrementally, as $\sumDist_{\cluster_p \cup \cluster_q} =
\sumDist_{\cluster_p} + \sumDist_{\cluster_q} + 2
\sumDist_{\cluster_p,\cluster_q}$.

\subsection{Fast Hierarchical Clustering Analysis}
Since Lance and Williams introduced the generalized recurrence formula in
1967, many works have been devoted to produce efficient algorithms for
agglomerative hierarchical clustering (see e.g.
\cite{murtagh85MultidimensionalClustering, rohlf73HierarchicalClustering}).
In this work, we rely on a very recent algorithm proposed by D. Müllner
\cite{Mullner11ModernHierarchical}.  Müllner's algorithm is a sophistication
of an older algorithm proposed by Anderberg \cite{Anderbeg73ClusterAnalysis}.
Even if the algorithm does not improve the worst case complexity of
Anderberg's algorithm ($O(N^3)$), it is much more efficient in practical
situations, as demonstrated by the author on various benchmarks
\cite{Mullner11ModernHierarchical}.  Moreover this algorithm can handle
general linkage criteria such as the one proposed in the previous section.

The efficiency of Müllner's algorithm can be explained succinctly as
follows. As the search for the closest pair of clusters in term of linkage is
the bottleneck of the naive implementation, Müllner's algorithm maintains a
priority queue in which a nearest neighbor cluster candidate is stored for
each cluster. The queue is sorted according to a lower bound of the distance
between each cluster and its true nearest neighbor. At each step of the
algorithm, the cluster of highest priority (smallest lower bound) is dequeued
from the structure, and the algorithm checks whether this cluster and its
candidate are indeed the closest pair (the distance between the cluster and
its candidate must be equal to the lower bound).  In such a case, the merge is
performed and the different data structures are updated.  Otherwise, the true
nearest neighbor is computed, the lower bound is replaced by the true distance
to the nearest neighbor, and the cluster's candidate is again inserted in the
priority queue.  This approach leads in practice to a very efficient search
for the closest pair of clusters. Additionally, it tends to delay  as
long as possible nearest neighbor computations.  As these searches are
postponed to latter steps in which the number of remaining clusters is
smaller, the algorithm is much more efficient.

\section{Multi-level refinement}

\subsection{Partition refinement}
Once the hierarchy is set up, the analyst is free to choose a partition by
cutting the dendrogram at a given level, using any adapted
heuristics. However, the partitions obtained this way are frequently
suboptimal. Indeed, as explained in e.g. \cite{Karypis99MultilevelRefinement},
during the construction of the hierarchy, bad merging decisions during the
early steps cannot be corrected in the later steps, which leads to a wrong
clustering solution.

A quite simple approach to improve a given partition is to rely on a greedy
refinement heuristic which performs a local search by moving objects from one
cluster to another. The simplest approach is known as the \emph{fast greedy
  heuristic}: the heuristic selects each object in turn; for a given object,
the cluster switch that leads to the largest decrease in the error measure $E$
is performed (all switches are considered). As long as moves which improve the
partition quality exist, the process continues.

This heuristic has the advantage to be quite simple and can be implemented
efficiently (calculations can usually be done incrementally). However, fast
greedy has an important drawback, it can easily be trapped in local minima as
only moves which decrease the error measure are allowed. 

\subsection{Multi-level refinement}
In the context of hierarchical clustering, heuristics described in the
previous section are known as single-level refinement approaches, as they
operate on just one level of the hierarchy: the bottom level.  Their main
property is that they only move one object (a singleton) at a time.  The
Multi-level refinement (MLR) approach, on the opposite, operates on different
levels of the hierarchy.  For an intermediate level, each move corresponds to
a displacement of several individuals (a cluster of the level in fact).  This
property allows the MLR to escape more easily from local minima than
single-level heuristics.  Indeed, for single-level heuristics, the
displacement of a dense group of items is unlikely to happen, as it would
imply many individual moves with a temporary large increase of the error
measure.  For the MLR, on the opposite, the displacement of a
group of items is done in just one operation, which avoids the increase of the
error measure.

Operating on all the levels of dendrogram would be too costly as it would
roughly multiply the cost of a refinement at a single level by $O(N)$. The MLR
is therefore built upon a selection of levels. Given a reduction factor
$0<\reductionFactor<1$, the MLR considers only the initial trivial partition
of singletons $\partition_1$ with $N$ clusters, and then a series of
partitions $(\partition_k)_{1\leq k\leq K}$ whose sizes are given by a
geometric progression based on $\reductionFactor$: $\partition_2$ contains
$\reductionFactor N$ clusters, $\partition_3$ $\reductionFactor^2 N$ clusters,
etc. $\partition_{K-1}$ is the last level before the one chosen by the analyst
(which corresponds to partition $\partition_{K}$).

Given the series of partitions $(\partition_k)_{1\leq k\leq K}$, the MLR
proceeds in a top-down way. As the partitions form a hierarchy, $\partition_{k}$
can always be considered as a partition of the clusters of
$\partition_{k-p}$ for any $p>0$. The main idea of MLR is to apply the fast
greedy heuristic to $\partition_K$ considered as a partition of
$\partition_k$ for $k$ decreasing from $K-1$ to $1$. More precisely, it first
refines $\partition_{K}$ considered as a partition of $\partition_{K-1}$: this
corresponds to moving entire sub-clusters of data points from one cluster of
$\partition_{K}$ to another one. Once this is done, the modifications of
$\partition_K$ are projected onto $\partition_{K-2}$ and the process
repeats. The final stage corresponds to applying fast greedy on the objects as
in the single level approach.

Notice that moving a sub-cluster $\cluster$ of a given partition
$\partition_k$ from its current cluster $\cluster_p$ to another cluster
$\cluster_q$ in the partition of interest $\partition_K$ leads to an increase
of the error measure:
\[
\Delta \error_{\cluster: p \rightarrow q} = 
\frac{\sumDist_{\cluster \cup \cluster_q}}{|\cluster|+|\cluster_q|}
- \frac{\sumDist_{\cluster_q}}{|\cluster_q|}
- \frac{\sumDist_{\cluster_p}}{|\cluster_p|}
+ \frac{\sumDist_{\cluster_p \setminus \cluster }}{|\cluster_p|-|\cluster|}
\]
Once again, sums $\sumDist_{\cluster \cup \cluster_q}$ and
$\sumDist_{\cluster_p \setminus \cluster }$ can be updated incrementally at
each move, leading to an efficient implementation.

\subsection{Related works}
Multi-level heuristics originate from graph clustering for which they give some
of the best results (see \cite{Hendrickson95MultilevelAlgorithm} for
the minimum cut partitioning problem, and \cite{Noack09MultilevelAlgorithms}
for the detection of communities in a network).  MLR has also been considered
for dissimilarity data in \cite{Karypis99MultilevelRefinement} as a way of
improving HCA results. In this paper, the authors extract a $k$-nearest
neighbour graphs from the dissimilarity matrix. They apply a standard HCA on
the graph using a variant of the average linkage criterion. Then they apply a
multi-level refinement approach to the hierarchical clustering using an error
measure close to the quantity $\error$ used here. Our proposal differs in
using the full dissimilarity matrix, in relying on the standard
quantization error for dissimilarity data in all the phases of the algorithm
and in leveraging Müllner's efficient HCA.

\section{Experiments}
The proposed method is tested on two classical dissimilarity data sets: the
small size cat cortex database with 65 objects (see e.g.,
\cite{GraepelObermayer1999DSOM}) and the large size Copenhagen chromosome
database with 4200 objects (see e.g. \cite{Neuhaus_Bunke_2006}). Reference
performances are provided by the relational k-means (RKM) in its standard (non
naive) implementation described in \cite{RossiEtAlFastRelationalWSOM2007}. In
all cases, the RKM is started from a number of initial random configurations
chosen so that both methods use approximately the same computational
ressources (on the same computer and using the same implementation language,
java). 

\begin{figure}[hbtp]
  \centering
    \includegraphics[width=0.8\linewidth]{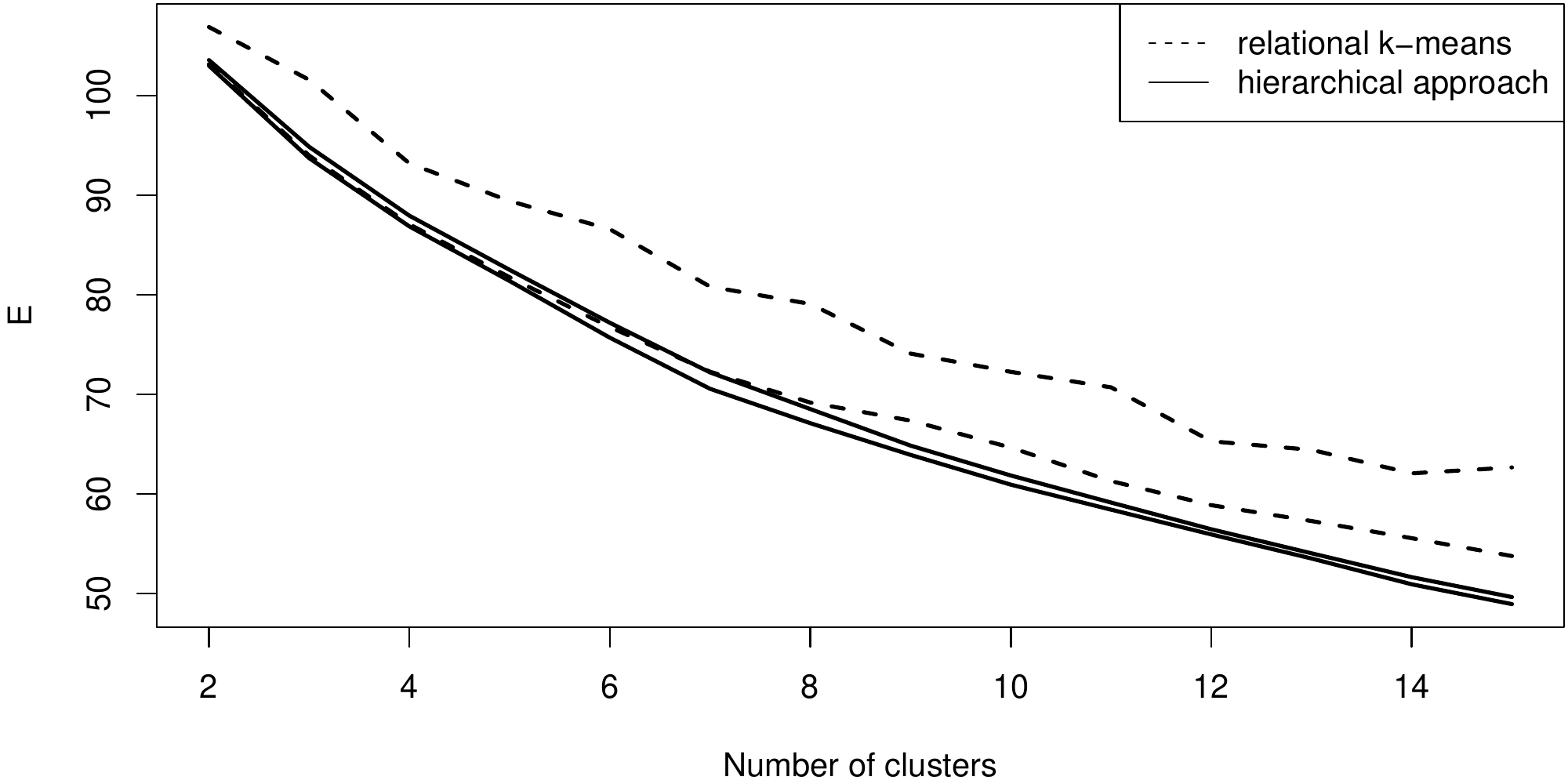}
\vskip -1em

  \caption{Quantization error $\error$ as a function of the number of clusters for the cat cortex database
(top solid line: standard HCA, bottom solid line: MLR, dashed lines: best and worst results of the RKM)}
\label{fig:cat}
\end{figure}

Results obtained on the cat cortex data set are summarized by Figure
\ref{fig:cat}.  For the relational k-means, the two dashed lines correspond to
the best and the worst results obtained out of 20 random initial
configurations. For the hierarchical approach, the top line corresponds to the
quantization error after the hierarchical phase while the bottom line shows
the error reached after the multi-level refinement. For the MLR, we set $\alpha=0.75$.
This value achieves a good balance between solution quality and MLR running time. 
The refinement does not bring much in this case (less than $2\%$ of decrease of
$E$), but it enables the hierarchical approach to beat the relational k-means
in all cases. It should be noted that even with 100 times more initial
configurations (a choice that increases significantly the running time of the
method), the RKM reaches the same quality as the hierarchical method only for
2 to 5 clusters.

\begin{figure}[hbtp]
  \centering
\includegraphics[width=0.8\linewidth]{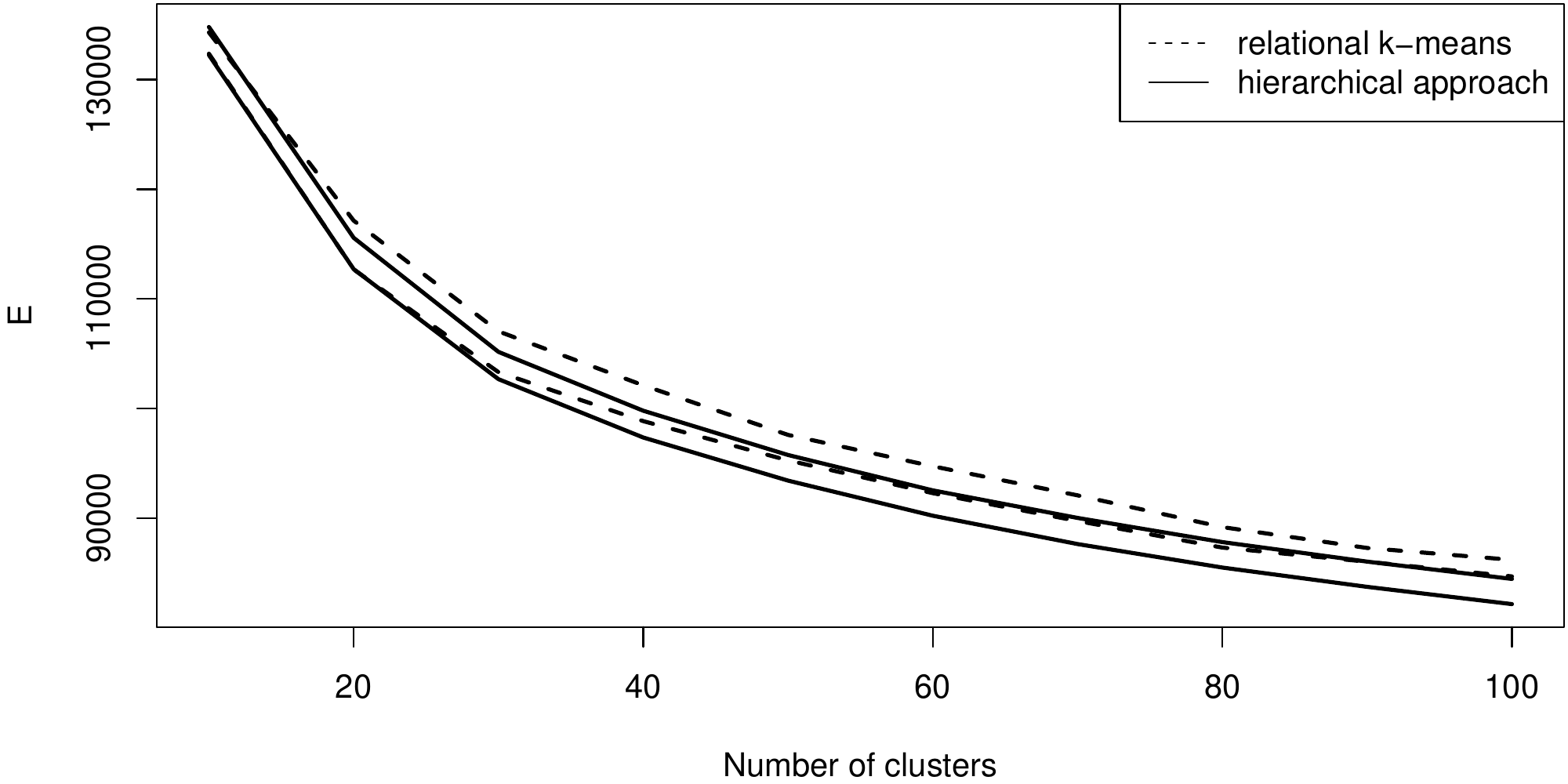}
\vskip -1em
\caption{Quantization error $\error$ as a function of the number of clusters for the chromosome database}
\label{fig:chrom}
\end{figure}

Similar results are obtained on the chromosome database (with only $10$
random initial configurations due to a higher computational load for the RKM
for this dataset), as shown on Figure \ref{fig:chrom}. Both methods perform
comparatively for a small number of clusters, while the hierarchical approach
outperforms the relational k-means when the number of clusters increases.

\section{Conclusion}
As shown in the experiments, the proposed hierarchical approach gives very
satisfactory results on real world datasets, especially when the number of
clusters chosen by the analyst is high. Further works will focus on exploiting
the full hierarchy as we have only used here the method in a k-means like
situation where only one clustering is considered.

\begin{footnotesize}

\bibliographystyle{abbrv}
\bibliography{hierarchical}

\end{footnotesize}


\end{document}